\documentclass[11pt,a4paper]{article}
\usepackage[hyperref]{emnlp2020}
\usepackage{times}
\usepackage{latexsym}

\usepackage{booktabs}

\usepackage{graphicx}
\usepackage{amsmath}
\usepackage{float}

\usepackage{microtype}
\usepackage{hyperref}
\usepackage{multirow}
\usepackage{color, colortbl}

\newcommand{\method}{\textsc{BERT-HiGRU}}
\date{}
\newcommand{\beginsupplement}{%
	\setcounter{table}{0}
	\renewcommand{\thetable}{S\arabic{table}}%
	\setcounter{figure}{0}
	\renewcommand{\thefigure}{S\arabic{figure}}%
}

\aclfinalcopy 
\def\aclpaperid{3072} 

\definecolor{LightCyan}{rgb}{0.88,1,1}

\newcommand{\Tref}[1]{Table~\ref{#1}}
\title{Keeping Up Appearances: Computational Modeling of Face Acts in Persuasion Oriented Discussions}

\author{Ritam Dutt \\\And
  Rishabh Joshi \\
 Language Technologies Institute\\
 Carnegie Mellon University\\
  \texttt{\{rdutt, rjoshi2, cprose\}@cs.cmu.edu} \\\And
  Carolyn Penstein Ros\'{e} \\ \\}

\date{}

\begin{document}
\maketitle
\begin{abstract}
The notion of \emph{face} refers to the public self-image of an individual that emerges both from the individual's own actions as well as from the interaction with others. 
Modeling face and understanding its state changes throughout a conversation is critical to the study of maintenance of basic human needs in and through interaction. Grounded in the politeness theory of \citet{brown1978universals}, we propose a generalized framework for modeling \emph{face acts} in persuasion conversations, resulting in a reliable coding manual, an annotated corpus, and computational models. The framework reveals insights about differences in face act utilization between asymmetric roles in persuasion conversations. Using computational models, we are able to successfully identify face acts as well as predict a key conversational outcome (e.g. donation success). Finally, we model a latent representation of the conversational state to analyze the impact of predicted face acts on the probability of a positive conversational outcome and observe several correlations that corroborate previous findings.

\end{abstract}

\section{Introduction}
Politeness principles, displayed in practice in day-to-day language usage, play a central role in shaping human interaction. Formulations of politeness principles are related to basic human needs that are jointly met in and through interaction \citep{grice1975logic,brown1987politeness,  leech2016principles}. Natural language offers various ways to enact politeness.  One of the most influential politeness theories from linguistics is proposed in \citep{brown1978universals}, in which a detailed exposition is offered of the individual actions whose cumulative effect results in \emph{saving face} and \emph{losing face}, along with a consideration of cost.  Using this framework, it is possible to analyze how interlocutors make decisions about where and how these devices should be used based on an intricate cost-benefit analysis \citep{brown1987politeness}. We refer to these component actions here as \textbf{face acts}.

The idea of \emph{face acts} appears quite attractive from a computational standpoint for their potential role in understanding what is ``meant'' from what is ``said'' \citep{grice1975logic,brown1987politeness,  leech2016principles}.
Consequently, politeness has been widely researched in various domains of language technologies \citep{walker1997improvising, gupta2007rude,wang2012love, abdul2012awatif,danescu-niculescu-mizil-etal-2013-computational} in addition to foundational work in pragmatics and sociolinguistics \citep{brown1987politeness, grice1975logic, leech2016principles}. However, much prior work modeling politeness reduces the problem to a rating task or binary prediction task, separating polite and impolite behavior, with the result that what is learned by the models is mainly overt markers of politeness or rudeness, rather than the underlying indirect strategies for achieving politeness or rudeness through raising or attacking face, even in the indirect case where no overt markers of rudeness or politeness might be explicitly displayed.


In contrast, the main contribution of this work is the investigation of eight major face acts, similar to dialogue acts, including an investigation of their usage in a publicly available corpus of \citet{persuasion-for-good-2019}.  In the selected corpus, a persuader (\textbf{ER}) is tasked with convincing a persuadee (\textbf{EE}) to donate money to a charity. The nature of the task prompts frequent utilization of face acts in interaction, and thus these face acts are abundantly present in the chosen dataset. We also provide a generalized framework for operationalizing face acts in conversations  as well as design an annotation scheme to instantiate these face acts in the context of persuasion conversations (\S\ref{sec:framework}, \S\ref{sec:annotation}). We offer the annotations we have added to this public dataset as another contribution of this work (\S\ref{sec:dataset}).
Additionally, we develop computational models to identify face acts  (\S\ref{sec:models-face-act-prediction}) as well as construct a latent representation of conversational state to analyze the impact of face acts on conversation success (\S\ref{sec:conversation-success}). We achieve  0.6 F1  on classifying face acts (\S\ref{sec:face_act_prediction}), and 0.67 F1 in predicting donation outcome (\S\ref{sec:outcome}). We observe that the predicted face acts  significantly impact the local probability of donation (\S\ref{sec:don_prob})\footnote{We include our annotation framework in Appendix and the annotated dataset and code is publicly available at \url{https://github.com/ShoRit/face-acts}}.




\section{Framework}
\subsection{Face Representation}
\label{sec:framework}
Face, based on the politeness theory of \citet{brown1987politeness}, reflects the `public self-image' that every rational adult member of society claims for himself. It can be subdivided into \emph{positive face}, referring to one's want to be accepted or valued by society, and \emph{negative face}, referring to one's right to freedom of action and freedom from imposition.

We refer to `face acts' as utterances/speech acts that alter the positive and/or the negative face of the participants in a conversation. We hereby refer to the acts that attack one's face as Face Threatening Acts (FTA) and those acts that raise one's face as Face Saving Acts (FSA). For example, criticizing an individual is an attack on the other's positive face, whereas refusing to comply with someone's wishes, raises one's own negative face. 
We also note that a single utterance or act can simultaneously affect the face of one or both participants in a conversation. For example, a refusal to donate to a charity because they do not trust the charity involves asserting one's negative face as well as decreasing the charity's positive face.

The implication of a face act between the participants is governed by several factors such as `power' and relative `social distance', as well as the relative threat (`ranking') of the face act \cite{brown1978universals}. For example, refusing to comply with the orders of one's superior is more threatening than requesting a friend for some change. 

Moreover, face acts need to be contextualized for a particular situation based on the rights and obligations of the individual participants, such as in compliance-gaining episodes \cite{compliance1988}. For example, a teacher has the responsibility and right to assign homework to the students. Such an action cannot be perceived as an attack on negative face, even though the student is reluctant to do so.

Based on the definition of face and face acts, we design a generalized annotation framework to capture the face dynamics in conversation. We instantiate our framework on a publicly-available corpus on persuasion dialogues.

\subsection{Dataset Description}
\label{sec:dataset}
We use the pre-existing persuasion corpus of \citet{persuasion-for-good-2019}. Each conversation comprises a series of exchanges where the persuader (ER) has to convince the persuadee (EE) to donate a part of their task earnings to the charity, \emph{Save the Children}. This selected corpus is well-situated for our task since each conversation is guaranteed to have a potential face threat (i.e., a request for money) and hence, we can expect face act exchanges between the two participants. It also sets itself apart from other goal-oriented conversations such as restaurant reservations and cab booking \cite{multi-domainwoz} since in those cases the hearer is obligated to address what might otherwise come across as a FTA (request/ booking), and thus in those cases non-compliance can be assumed to be due to logistic issues rather than an unwillingness to co-operate. 

In the selected corpus, the participants are Amazon Mechanical Turk workers who are anonymous to each other, which controls for the `social distance' variable. Moreover, the participants have similar `power', with one role having some appearance of authority in that it represents an organization, but the other role representing possession of some desired object (i.e., money). Thus, we argue that although ER imposes an FTA by asking for donation, EE is equally at liberty to refuse. Moreover, ER does not incur a penalty for failing to persuade. In fact the corpus includes some conversations that do not talk about donation at all. We also emphasize that the task was set up in a manner such that EE come into the interaction blind to the fact that ER have been tasked with asking them to donate.

We assess the success of a conversation based on whether EE agrees to donate to the charity. We label successful conversations as \emph{donor} conversations and \emph{non-donor} conversation otherwise.  We refer the reader to \citet{persuasion-for-good-2019} for more details about the dataset.

\subsection{Annotation Framework}
\label{sec:annotation}

\begin{table*}[h]
\small
\centering

\begin{tabular}{p{0.07\textwidth} p{0.85\textwidth}}
\toprule
Face Act & Description \\ \midrule
SPos+ & (i) S posit that they are \textbf{virtuous} in some aspects  or they are \textbf{good}. \\
&(ii) S \textbf{compliment the brand or item {they represent or endorse}} and thus project their credibility.\\
&(iii) S state their \textbf{preference or want}, something that they like or value.\\ \hline

SPos-& (i) S \textbf{confess} or apologize for being unable to do something that is expected of them.  \\
& (ii) S \textbf{criticise or humiliate themselves}. They damage their reputation or values by either saying they are not so virtuous or criticizes some aspect of the brand/item they endorse or support.\\ \hline

HPos+& (i) \textbf{S compliment H} either for H's virtues, efforts, likes or desires.  It also extends to S \textbf{acknowledging the efforts of H} and \textbf{showing support for H}.\\ 
& (ii) S can also provide an \textbf{implicit compliment to incentivize H }to do something good.\\
& (iii) S \textbf{empathize} / sympathize or in general \textbf{agree with H}.\\
& (iv) S is \textbf{willing to do the FTA }as imposed by H (implying that the FTA is agreeable to S.)\\ \hline
HPos-&(i) S voice \textbf{doubts or criticize H }or the product/brand that H endorses.\\
& (ii) S \textbf{disagree with H }over some stance, basically contradicting their viewpoint. \\
&(iii) S is either \textbf{unaware} or \textbf{indifferent} to H's wants or preferences. \\   \midrule \midrule
SNeg+& 
(i) S \textbf{reject} or are unwilling to do the \textbf{FTA.} Stating the reason does not change the circumstances of non-compliance but sometimes helps to mitigate the face act.\\ \hline
SNeg-&  (i) S \textbf{offer to assist H}.\\ \hline
HNeg+ &
(i) S seek to \textbf{decrease the imposition of the FTA} on H by either decreasing the inconvenience such as providing alternate, simpler ways to carry out the FTA or decrease the threat associated with the FTA. \\
& (ii) S \textbf{apologize for the FTA} to show that S understood the inconvenience of imposing the request but they have to request nevertheless. \\ \hline
HNeg-&
(i) S \textbf{impose an FTA on the H}. The FTA is some act which H would not have done on their own. \\
&(ii) S \textbf{increase the threat} or ranking of the FTA \\
&(iii) S \textbf{ask/request H} for assistance?\\
\bottomrule
\end{tabular}
\caption{Generalized framework for situating and operationalizing face acts in conversations. The predicates for each of the face act are highlighted in bold.}
\vspace{-0.3cm}
\label{tab:face-acts}
\end{table*}

In a two-party conversation, a face act can either raise (+) or attack (-) the positive face (Pos) or negative face (Neg) of either the speaker (S) or the hearer (H), leading to 8 possible different outcomes. For example, \textit{HPos+} means raising the positive face of the hearer. We provide a generalized framework in Table \ref{tab:face-acts} for labelling a speech act / utterance with one or more face acts, building upon the politeness theory of \citet{brown1978universals}. The framework is designed to be explicit enough to ensure the creation of a reliable coding manual for classifying face-acts, as opposed to the simple classification of requests and other directives as intrinsic FTAs \cite{brown1978universals}. Moreover, since we also seek to operationalize FSA, we make some departure from the original classification of directives. For example, we feel that compliments  directed at the hearer, should  be \textit{HPos+}  rather than \textit{HNeg-} (as observed in \citet{brown1987politeness}) 
since an appreciation for someone’s efforts is more desirable.

We highlight the predicates that result in a particular face act in bold in Table \ref{tab:face-acts}. For example, S claiming to be virtuous or doing some good deed amounts to raising their own positive face (\textit{SPos+}). Although the framework is designed to be generalizable across domains, the predicates themselves need to be instantiated based on the domain of choice. For example, in this particular corpus, the act of requesting someone for donation counts as a FTA. We refer the readers to Table \ref{tab:predicates } in Appendix \ref{sec:appendix} which outlines how the face acts are instantiated for the specific persuasion dataset.

Each conversation in the dataset consists of $10$ or more turns per participant with one or more utterances per turn. Each utterance is labeled with one or more face acts according to our annotation framework, or `Other' if no face act can be identified, or if the utterance contains no task-specific information (Eg: Small talk). We consider ER to be a representative of the charity since ER advocates for donations on their behalf. We show the flowchart detailing the annotation framework in Figure \ref{fig:my_label} of Appendix \ref{sec:appendix}. 

\noindent{\textbf{Validating the annotation scheme:}} Two authors of the paper annotated $296$ conversations in total. 
The annotation scheme underwent five revisions, each time with three different conversations, eventually yielding a high Cohen's Kappa score of $0.85$ across all face acts \cite{cohen1960coefficient}. 
The revised scheme was then used to annotate the remaining conversations. 
We show an annotated conversation snippet in \Tref{Table: Annotated Conversation Snippet}.

    
    

\subsection{Summary Statistics}
\label{sec:pa}

Our annotated dataset comprises $231$ donor conversations and $65$ non-donor conversations. \Tref{Table: Dataset Stats} shows the distributions of different face acts employed by ER and EE respectively for both donor and non-donor conversations. We also note that certain face-acts do not occur in our corpus, such as \textit{SPos-} for ER, presumably because ER does not have a reason to debase themselves or the charity they endorse. We provide a detailed explanation of the occurrence of such acts in the supplementary section. 
We observe multiple statistically significant differences in face act prevalence based on whether EE is a donor or non-donor. 
Some findings are intuitive, such as an increase in \textit{HPos+} for Donor conversations (for both ER and EE). We argue that EE had acknowledged the efforts of the charity and was willing to donate, and was thus rewarded with compliments from ER. Likewise, \textit{SNeg+} occurs significantly more in Non-donor situations, due to a refusal to comply. We note that a majority of the turns labeled `Other' involve greetings or conversation exchanges unrelated to the main business of the conversations.



\begin{table}[!htbp]
\centering
\small
\begin{tabular}{lllll}
\toprule
Face Acts &  \multicolumn{2}{c}{ER} & \multicolumn{2}{c}{EE}\\
\midrule
{}   & \multicolumn{1}{l}{{ }{ }{ }D}   & \multicolumn{1}{l}{{ }{ }{ }N}    & \multicolumn{1}{l}{{ }{ }{ }D}  &  \multicolumn{1}{l}{{ }{ }{ }N}\\
SPos+ & 19.95& 23.03 &  { }8.29&  {  }6.51\\
SPos- & { }0.00 & { }0.00 &  { }0.18& $ { }{ }{ }0.96^{*}$\\
HPos+ & $23.08^{***}$ & 16.24& $36.17^{***}$& 21.07\\
HPos- &  { }0.70 & { }$2.65^{*}$ &  { }4.37 & $10.73^{**}$ \\
SNeg+ &  { }0.00 &  { }0.00 &  { }3.85 & $11.97^{***}$\\
HNeg+ &  { }5.50&  {  }4.81&  { }0.00 &  { }0.00\\
HNeg- &10.47& 10.85 &  { }9.20 & 13.03\\
Other & 40.31& 42.42 & 37.94 & 35.73\\
\bottomrule
\end{tabular}
\caption{Distribution of different face acts for the donor (D) and  non-donor (N) for ER and EE. *, **, and *** signify that the specific act is statistically significant for D and N according to the independent t-test with p-values $\leq$ $0.05$, $0.01$, and $0.001$ respectively.}
\label{Table: Dataset Stats}
\end{table}
\vspace{-0.5cm}

\begin{figure*}[h]
    \centering
    \includegraphics[scale =0.22]{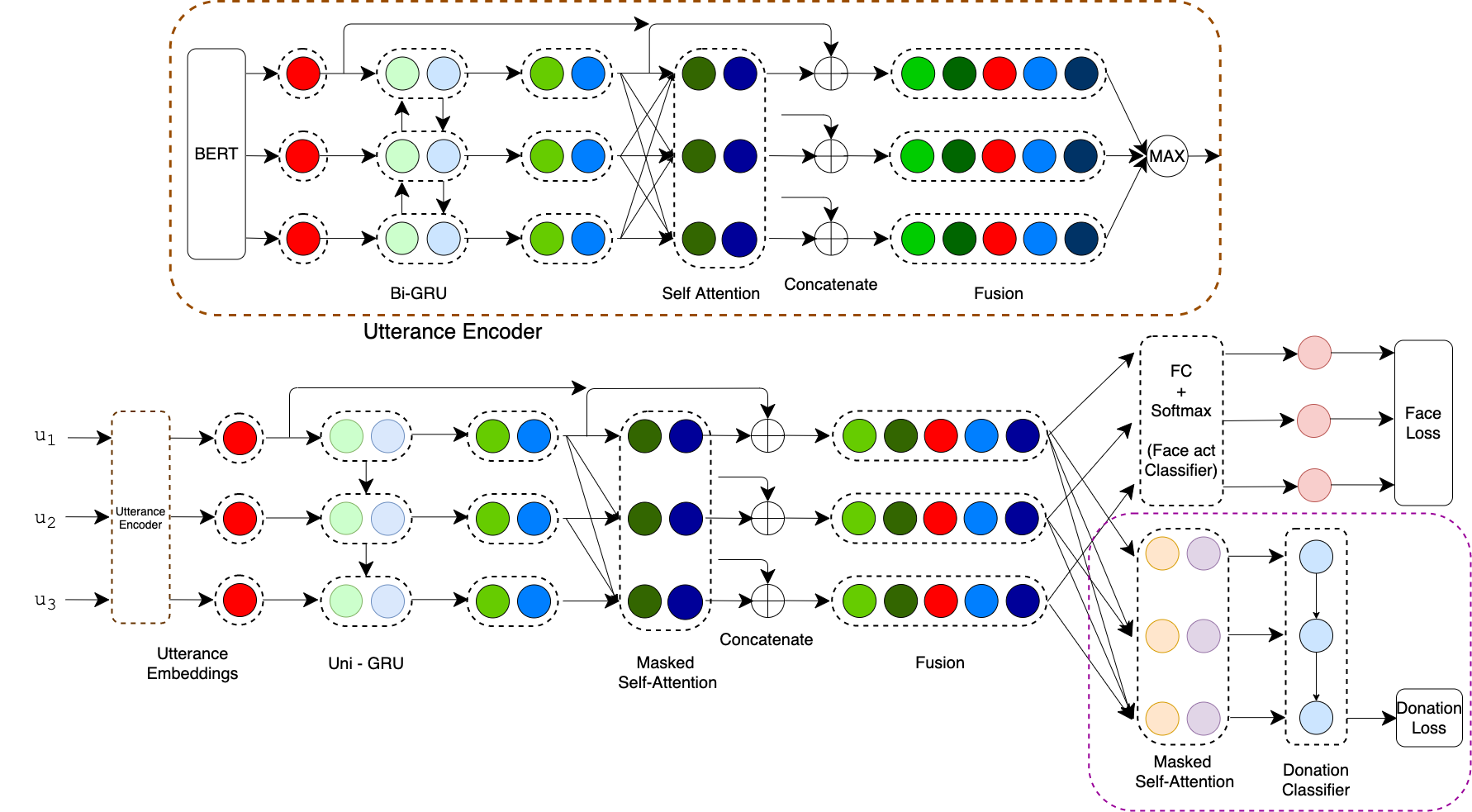}
    \caption{Overview of \method{}. We first encode the utterances by passing the BERT representations of the token through a BiGRU layer followed by Self Attention. The BERT, BiGRU and Self-Attention outputs are then fused to get the final token representation before max pooling. This utterance representation is passed through a uni-directional GRU followed by Masked-Self-Attention and fusion. One part of the model uses the face classifier to predict the face-act of each utterance while the other model uses another layer of Masked-Self-Attention to predict the donation probability. The details are in Section \ref{sec:models}}
    \label{fig:my_label}
    \vspace{-0.6cm}
\end{figure*}

\section{Methodology}
\label{sec:models}
\subsection{Face act prediction}
\label{sec:models-face-act-prediction}
We model the task of computationally operationalizing face acts as a dialogue act classification task. Given an dialogue with $n$ utterances, $D = [u_{1},u_{2}, ..., u_{n}]$, we assign labels $y_{1}, y_{2} ... y_{n}$ where $y_{i} \in Y$ represents one of 8 possible face acts or `Other'. Although, we acknowledge that an utterance can have multiple face acts, we observe that multi-labeled utterances comprise only $2\%$ of our dataset, and thus adopt the simplification of predicting a single face-act for each utterance\footnote{For each utterance with multiple labels, one is randomly select from that set to be treated as the Gold label.}. Several tasks in the dialogue domain, such as emotion recognition \cite{majumder2019dialoguernn, HiGRU}, dialogue act prediction \cite{chen2018DAP,raheja-tetreault-2019-dialogue} and open domain chit-chat \cite{persona_chat, amused}, have achieved state-of-the-art results using a hierarchical sequence labelling framework. Consequently, we also adopt a modified hierarchical neural network architecture of \citet{HiGRU} that leverages both the contextualized utterance embedding and the previous conversational context for classification. We hereby adopt this as the foundation architecture for our work and refer to our instantiation of the architecture as \method{}. 

\noindent{\textbf{Architecture of \method{}:}} An utterance $u_{j}$ is composed of tokens $[w_0, w_1, ..., w_K]$, which are represented by their corresponding embeddings $[e(w_0), e(w_1),..., e(w_K)]$. In \method{}, we obtain these using a pre-trained BERT model \cite{devlin2019bert}.
We pass these contextualized word representations through a BiGRU to obtain the forward $\overrightarrow{h_{k}}$ and backward $\overleftarrow{h_{k}}$ hidden states of each word, before passing them into a Self-Attention layer. This gives us corresponding attention outputs, $\overrightarrow{ah_{k}}$ and $\overleftarrow{ah_{k}}$. Finally, we concatenate the contextualized word embedding with the GRU hidden states and Attention outputs in our \textit{fusion layer} to obtain the final representation of the word. We perform max-pooling over the fused word embeddings to obtain  the $j^{th}$ utterance embedding, $e(u_j)$. Formally, 
\begin{equation*}
    \begin{split}
       \overrightarrow{h_{k}} =& \operatorname{GRU}\left(e\left(w_{k}\right), \overrightarrow{h_{k-1}}\right) \\
       \overleftarrow{h_{k}} =&\operatorname{GRU}\left(e\left(w_{k}\right), \overleftarrow{h_{k+1}}\right)\\
       \overrightarrow{ah_{k}} =&\operatorname{SelfAttention}(\overrightarrow{h_{k}})\\
       \overleftarrow{ah_{k}} =& \operatorname{SelfAttention}(\overleftarrow{h_{k}})\\
       e_c(w_k) = \operatorname{tanh}(&W_w[\overrightarrow{ah_{k}}; \overrightarrow{h_{k}}; e(w_k); \overleftarrow{h_{k}}; \overleftarrow{ah_{k}}] + b_w)
    \end{split}
\end{equation*}
\vspace{-5mm}
\begin{equation}
\label{eqn:one}
e(u_j) = \operatorname{max}(e_c(w_1), e_c(w_2),... e_c(w_K))
\end{equation}
Similarly, we calculate the contextualized representation of an utterance $e_c(u_j)$ using the conversation context. 
In departure from \citet{HiGRU}, we pass $e(u_{j})$ through a uni-directional GRU that yields 
the forward hidden state $\overrightarrow{H_{j}}$. Masked Self-Attention over the previous hidden states, yields $\overrightarrow{AH_{j}}$. We fuse $e(u_j)$, $\overrightarrow{H_{j}}$ and  $\overrightarrow{AH_{j}}$ before passing it through a linear layer with tanh activation to obtain  $e_c(u_j)$. This ensures that current $e_c(u_j)$ is not influenced by future utterances, enabling us to observe change in donation probability over time in Section \ref{sec:conversation-success}
\begin{equation*}
    \begin{split}
        \overrightarrow{H_{j}} =& \operatorname{GRU}\left(e\left(u_{j}\right), \overrightarrow{H_{j-1}}\right)\\
        \overrightarrow{AH_{j}} =& \operatorname{MaskSelfAttention}(\overrightarrow{H_{j}})    
    \end{split}
\end{equation*}
\vspace{-0.2cm}
\begin{equation}
    \label{eqn:two}
    e_c(u_j) = \operatorname{tanh}(W_u[\overrightarrow{AH_{j}}; \overrightarrow{H_{j}}; e(u_j)] + b_u)
\end{equation}


We explore different hierarchical architecture variants which differ in the creation of contexualized embeddings $e_c(w_k)$ and $e_c(u_j)$ in Equation \ref{eqn:one} and \ref{eqn:two}. (1) \method~ includes only the final hidden state $\overrightarrow{H_{j}}$; (2) \method{}-f additionally employs the utterance embedding $e_(u_j)$; and (3) \method{}-sf, which also includes the attention vector $\overrightarrow{AH_{j}}$. 



We feed the final contextualized utterance embedding $e_c(u_j)$ through a FC layer with dropout and project it onto the state space of face-acts. We then apply softmax to obtain a probability distribution over the face-acts, with negative logarithmic loss as the loss function. Given the true labels $y$ and the predicted labels $y'$, the loss is computed for all $n$ utterances in a conversation as:

\vspace{-0.3cm}
\begin{equation}
    L_{f} = -\sum_{i=1}^{n} \sum_{y_j \epsilon Y}y_jlog(y'_{j})
\end{equation}
\vspace{-0.3cm}

\subsection{Impact of face acts on donation}
\label{sec:conversation-success}
\textbf{Donation Outcome Prediction:} \citet{brown1987politeness} notes that the exchange of face acts contributes towards an evolving conversational state. We seek to view the evolving state representation within our sequence model and analyze its impact on conversation success. The best reflection of what the evolving conversational state accomplishes in the context of persuasion is whether a donation occurs or not.  We thus add the prediction task as supervision and interpret the resulting conversation state based on how the probability of donation changes. We accomplish the supervision by incorporating another loss, called donation loss, in addition to the loss obtained for face acts.  

For each utterance $u_{j}$, we apply masked self attention over the set of contextual utterance embeddings  $e_c(u_j)$ till the $j^{th}$ utterance and project it through a linear layer with tanh activation to obtain the donation score $don_{j}$. The tanh non-linearity ensures that the donation score remains between -1 and 1 and intuitively denotes the delta change in scores from the previous step. We finally compute the probability of donation $o'_{j}$ at the $j^{th}$ step, by applying sigmoid activation over the sum of probability at the previous step and the delta change $don_j$. This ensures that the $o'^{th}_{j}$ probability is restricted between 0 and 1. We obtain the donation loss $L_{d}$ similar to \citet{diyi-yang-persuasive} by taking the mean squared error  of the donation probability at the last step $o'_{n}$ and the actual donation outcome $o_{n}$. $o_{n}$ is 1 if successful, otherwise 0. We also experiment with Binary Cross Entropy loss and obtain similar results.
\begin{equation*}
    \begin{split}
         e_d(u_j) =& \operatorname{MaskSelfAttention}(e_c(u_j))\\
         don_j  =& \operatorname{tanh}(W_d[e_d(u_j)]+ b_d)\\
         o'_j  =& \sigma(o'_{j-1}+don_j)\\
         L_{d} =& (o'_{n}- o_n)^{2}
    \end{split}
\end{equation*}

The donation loss is combined with the original face-act loss in a weighted fashion using some hyperparameter $\alpha$, such that $\alpha$ $\epsilon$  $[0,1]$.
\begin{equation}
    \label{eqn:loss}
    \vspace{-0.1cm}
    L_{tot} = \alpha L_{f} + (1-\alpha)L_{d}
    \vspace{-0.1cm}
\end{equation}

\noindent \textbf{Correlating face acts with donation outcome:}
The aforementioned formulation enables us to obtain the donation probability at any given step and assess the impact of difference in conversational state (due to a specific face act) on the local assessment of the probability of donation. To  quantify the impact, we perform linear regression with the donation probability at each time step ($y_{i}$) as the dependent variable. The independent variables includes the predicted face acts for that step $(f_{i}^{k})$ and the donation probability at the previous step $y_{i-1}$. 
\begin{equation}
    \label{eqn:regression}
    y_{i} = \beta_{0}*y_{i-1} + \sum_{f^{k}}\beta_{k}*f_{i}^{k}
\end{equation}
Here, $\beta_{k}$ represents the coefficient of the corresponding face-act and $\beta_{0}$ the coefficient for $y_{i-1}$.

\section{Experimental Setup}
We describe the baselines and evaluation metrics here. We present the additional experimental details of our model in Appendix Table \ref{tab:hyperparameters}.
\subsection{Baselines}

\noindent\textbf{Face act prediction:}
We employ different variants of \method{} described in Section \ref{sec:models}, namely the vanilla \method{}, \method{}-f (with residual connections (fusion)) and \method{}-sf with self-attention and fusion.

To observe the effect of incorporating conversation context, we pass the utterance embedding $e(u_j)$ obtained from the utterance encoder directly into the face act classifier. We denote the different variants of utterance encoder employed as BERT-BiGRU, BERT-BiGRU-f, and BERT-BiGRU-sf with the same notation as the hierarchical variants. 

To explore the impact of embedding choice on model performance, we experiment with pre-trained Glove embeddings \cite{pennington2014glove} in addition to BERT tokens. We denote the hierarchical models with GloVe embeddings as HiGRU and those without conversation context as BiGRU.

\noindent\textbf{Donation Outcome Prediction:}
We use the baselines mentioned above for predicting face acts and augment them with the donation loss component. We explore different values of $\alpha$ for weighing the two losses as described in Equation \ref{eqn:loss}.

\subsection{Evaluation Metrics}
\noindent\textbf{Face act prediction:} We observe the model performance in predicting the face acts for (i) only the persuader (ER), (ii) only the persuadee (EE), and (iii) both ER and EE (All). We perform five-fold cross-validation due to the paucity of annotated data. We report performance in terms of mean accuracy as well as macro F1-scores across the five folds due to the high class imbalance.

\noindent\textbf{Donation Outcome Prediction:} For a given conversation, we observe the probability of donation at the final step as $o'_{n}$.  We choose an appropriate threshold on $o'_{n}$ across the five validation folds, such that a conversation with ($o'_{n}$) greater than the threshold is considered successful and vice versa. The F1 score is then computed on the binarized outcome. We choose macro F1-score as the evaluation metric due to the highly skewed distribution of the number of donor and non-donor conversations.

\noindent\textbf{Correlating face acts with donation outcome:} We quantify the impact of face acts on donation probability for both ER and EE through the corresponding coefficients obtained from the regression framework. 
A positive coefficient implies that the face act is positively correlated with the local donation probability and vice versa. We also note the fraction of times a particular face act contributed to a local increase in donation probability and denote it by \textit{Frac}. A value of \textit{Frac} $>0.5$ for an act implies that the act increased the local donation probability more number of times than it decreased it. 

\section{Results}

In this section we put forward the following research questions and attempt to answer the same.
\begin{itemize}
\item[Q1.] How well does \method{} predict face acts? (Section \ref{sec:face_act_prediction})	
\item[Q2.] How well are we able to predict the outcome of the conversation? (Section \ref{sec:outcome})
\item[Q3.] How do individual face acts correlate with donation probability? (Section \ref{sec:don_prob})
\end{itemize}	


\subsection{Face Act Prediction}
\label{sec:face_act_prediction}

\noindent{\textbf{Model Performance:}} We present the results of our models for face act prediction in Table \ref{Table: Model Results} and glean several insights. Firstly, we observe that all models consistently perform better for ER than EE due to the more skewed distribution of EE and the presence of an extra face act (6 for ER vs 7 for EE). The difference in F1 is less noticeable between EE and All, despite an additional face act (8 for All), possibly due to the increase in labelled data. 


\begin{table}[h]
\centering
\resizebox{7.5cm}{!}{\begin{tabular}{lcccccc}
\toprule
 Model &  \multicolumn{2}{c}{ER} & \multicolumn{2}{c}{EE} &  \multicolumn{2}{c}{All} \\
\midrule
{}   & Acc   & F1    & Acc   & F1 & Acc   & F1\\
BiGRU	 &0.68	&0.58	&0.59	&0.49	&0.62	&0.49\\
BiGRU-f	 &0.69	&0.57	&0.59	&0.50	&0.62   &0.49\\	
BiGRU-sf &0.69	&0.57	&0.60	&0.51	&0.62	&0.51\\ \hline
HiGRU    &0.68  &0.56	&0.62	&0.52	&0.63	&0.49\\
HiGRU-f	 &0.70	&0.57	&0.62	&0.52	&0.64	&0.49\\
HiGRU-sf &0.69	&0.59	&0.62	&0.56	&0.64	&0.52\\ \hline
BERT-BiGRU   	&0.73	&0.63	&0.66	&0.57	&0.67	&0.56\\
BERT-BiGRU-f 	&0.72	&0.62	&0.66	&0.58	&0.66	&0.57\\
BERT-BiGRU-sf	&0.72	&0.62	&0.65	&0.56	&0.66	&0.56\\
\hline
BERT-HiGRU & \textbf{0.74} &\textbf{0.63} & 0.66& 0.54 & 0.68&0.57\\
BERT-HiGRU-f & 0.73 & \textbf{0.63 }& \textbf{0.67}&\textbf{0.61} & \textbf{0.69} & \textbf{0.60}\\
BERT-HiGRU-sf & 0.73 & 0.62 & \textbf{0.67}& 0.60 & 0.68&0.59\\


\bottomrule
\end{tabular}}
\caption{Performance of the various models on face act prediction. The best results are shown in bold.}
\label{Table: Model Results}
\vspace{-0.3cm}
\end{table}

Adding conversational context aids model performance as observed by the average increase of 1.3 in F1 scores across all model configurations (BiGRU to HiGRU). The highest gains are observed for \method{}-sf and HiGRU-sf which attends over the set of previous utterances and thus can better reason about the current utterance. 

The greatest boost, however, comes from incorporating BERT embeddings as opposed to pre-trained GloVe, which bears testimony to the efficacy of contextualized embeddings for several dialogue tasks \cite{yu2019midas,lai2019simple}.

In fact, with the inclusion of BERT, \method{}-f performs as competitively as \method-sf. We also note that the performance of \method{}-f and \method{}-sf are statistically significant as compared to the other baselines according to the McNemar's test \cite{mcnemar1947note} with p-value $\leq$ 0.001.



\begin{figure}[t]
    \includegraphics[scale=0.5]{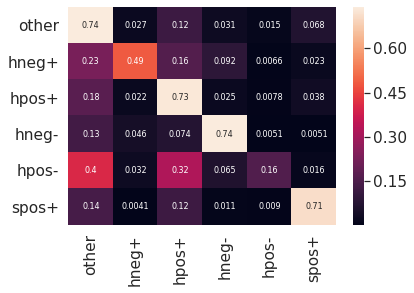}
    \caption{Confusion matrix for Persuader (ER)}
    \label{fig:ER-cm.}
    
    \includegraphics[scale=0.5]{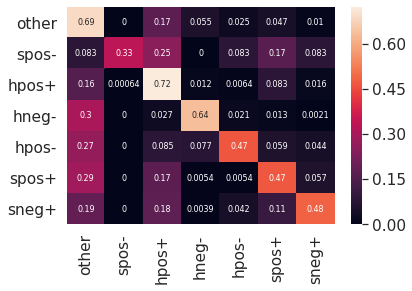}
    \caption{Confusion matrix for Persuadee (EE)}
    \label{fig:EE-cm.}
    \vspace{-0.5cm}
\end{figure}

\noindent\textbf{Error Analysis: } We present the confusion matrix of face act prediction for the \method{}-f model for both ER and EE in Figures \ref{fig:ER-cm.} and \ref{fig:EE-cm.} respectively. We observe that a majority of the misclassification errors occurs when a face-act is predicted as `Other' since it is the most frequent class and also shares a commonality with the remaining classes.


Some instances of misclassification errors for ER involves the labels \textit{HPos+} and \textit{HNeg+}. For example, utterances like  ``Every little bit will help the children we are trying to save.'' or ``Anyways, even 50 cents goes a long way you know?'' seek to decrease the imposition of the face threat by letting EE choose the donation amount, but also simultaneously provides EE with an opportunity to raise their face.  For EE, the face act \textit{SNeg+} is frequently misclassified as \textit{ SPos+} or \textit{HPos+}, since a refusal to donate is often accompanied by EE stating their preference  (``there are other charities I'd like to donate to over this one'') or acknowledging the efforts of ER and the charity (``I appreciate you informing about this but I'll be donating at a later time''). The greatest misclassification happens for \textit{HPos-} for ER and \textit{SPos-} for EE, since they account for $\approx$ 1\% and $\approx$ 0.3\% of the dataset respectively.  Consequently, they are often misclassified as \textit{HPos+} and \textit{SPos+} respectively since they affect the same face.


\begin{table*}[t]
\footnotesize
\centering
\resizebox{\textwidth}{!}{
\begin{tabular}{llccc}
\toprule
Role &  Utterance & True Face & Pred Face & Don Prob \\
\midrule
ER &	Hi	& other	& other	& 0.268 \\
\rowcolor{LightCyan}
EE &	hello	& other	& other	& 0.324 \\
ER &	How's it going	& other	& other	& 0.337 \\
\rowcolor{LightCyan}
EE &	it's going good so far this morning for me.	& other	& other	& 0.340 \\
\rowcolor{LightCyan}
EE &	what about you?	& other	& other	& 0.340 \\
ER &	Another day in paradise	& other	& other	& 0.340 \\
\rowcolor{LightCyan}
EE &	haha.	& other	& other	& 0.340 \\
\rowcolor{LightCyan}
EE &	do you live in the states ?	& other	& other	& 0.791 \\
ER &	Great state of Texas.	& other	& other	& 0.448 \\
ER &	How about you?	& other	& other	& 0.365 \\
\rowcolor{LightCyan}
EE &	chicago.	& other	& other	& 0.346 \\
\rowcolor{LightCyan}
EE &	so what is this about anyway?	& other	& other	& 0.792 \\
\rowcolor{LightCyan}
EE &	it says a child's charity.	& other	& other	& 0.448 \\
\rowcolor{LightCyan}
EE &	I am lost.	& other	& other	& 0.365 \\
ER &	I guess I'm supposed to persuade you to donate your bonus to the save the children charity.....	& hneg-	& other	& 0.346 \\
ER &	I want you to keep your bonus obviously	& hneg-	& other	& 0.342 \\
EE &	it would help to know what childrens charity as well with some information about it.	& hpos+	& other	& 0.341 \\
\rowcolor{LightCyan}
EE &	usually thats how you persuade ppl to donate.	& other	& other	& 0.341 \\
\rowcolor{LightCyan}
EE &	show a pic of a kid etc.	& other	& other	& 0.340 \\
ER &	How much do you like to donate to the charity now?	& hneg-	& hneg-	& 0.340 \\
ER &	Your donation will be directly deducted from your task payment.	& hneg+	& other	& 0.791 \\
ER &	You can choose any amount from \$0 to all your payment	& hneg+	& other	& 0.856 \\
\rowcolor{LightCyan}
EE &	I do not wish to donate.	& sneg+	& sneg+	& 0.464 \\
\rowcolor{LightCyan}
EE &	I've been given no info about the charity.	& hpos-	& sneg+	& 0.369 \\
ER &	They help children in warzones and other poor nations to get food and clothes	& spos+	& spos+	& 0.797 \\
\rowcolor{LightCyan}
EE &	oh ok, who is they?	& hneg-	& hneg-	& 0.449 \\
\rowcolor{LightCyan}
EE &	what is the organization ?	& hneg-	& hneg-	& 0.365 \\
ER &	Save The Children is the name	& spos+	& other	& 0.796 \\
\rowcolor{LightCyan}
EE &	no, I do not wish to donate at this time.	& sneg+	& sneg+	& 0.449 \\
\rowcolor{LightCyan}
EE &	there are other charities I'd like to donate to over this one.	& spos+	& sneg+	& 0.802 \\
\rowcolor{LightCyan}
EE &	I'm sorry.	& hpos+	& spos-	& 0.858 \\
\rowcolor{LightCyan}
EE &	I don't have a lot to work with either.	& sneg+	& sneg+	& 0.464 \\
ER &	Think of the poor kids in Syria who could get so much for the price of a coffee in Chicago.	& other	& hpos+	& 0.812 \\
ER &	Do you really need all you have when they have nothing at all?	& hpos-	& other	& 0.453 \\
\rowcolor{LightCyan}
EE &	I don't have much money for myself either which is why I consider this to be my part time job.	& sneg+	& sneg+	& 0.366 \\
\rowcolor{LightCyan}
EE &	I already work full-time to make ends meet.	& sneg+	& sneg+	& 0.346 \\
\rowcolor{LightCyan}
EE &	I'm sorry	& hpos+	& spos-	& 0.793 \\
ER &	But you have a full time job, food, shelter and I'm sure you have family and friends.	& hpos+	& other	& 0.857 \\
ER &	These kids families were murdered and they live in rubble that used to be their home, & & & \\
 & nobody to care for them and they only eat what they can find off the street from the dead.	& other	& hpos+	& 0.464 \\
ER &	If they eat at all	& hpos-	& hpos+	& 0.369 \\
\rowcolor{LightCyan}
EE &	that is very sad.	& hpos+	& hpos+	& 0.797 \\
\rowcolor{LightCyan}
EE &	I'd like to look into this charity more before I donate as well.	& sneg+	& other	& 0.857 \\
\rowcolor{LightCyan}
EE &	I'd like to see how the money is dispersed in the company	& hneg-	& other	& 0.464 \\
    \bottomrule
    \end{tabular}}
    \caption{An example conversation consisting of true and predicted face acts, along with donation probabilities. The persuader was unsuccessful in convincing the EE to donate. For brevity, the utterances of the EE are in cyan.}
    \label{Table: Annotated Conversation Snippet}
\vspace{-0.4cm}
\end{table*}

\subsection{Donation Outcome Prediction}
\label{sec:outcome}
We use \method{}-f as the model since it achieves the highest performance on face act prediction. Experiments with different values of alpha reveal that $\alpha = $ 0.75 achieves the highest F1 score on both face acts (0.591) and donation outcome (0.672).  The F1-score for face acts is comparable to the best performing model in \ref{sec:face_act_prediction} but the performance for donation outcome increases significantly from 0.545 to 0.672.  When $\alpha$ = 1.0, the donation outcome prediction is similar to random choice. On the other extreme, when $\alpha$ = 0, i.e. in the absence of $L_{f}$, the donation outcome behaves like random choice since, the model is unable to learn an appropriate latent state representation. 

\subsection{Correlation of face acts with donation.}
\label{sec:don_prob}
We present the coefficients of face acts for ER and EE obtained from the regression framework \ref{eqn:regression} and the corresponding fraction (Frac) of times the face act increased the donation probability in Table \ref{tab:correlation-donation-face-acts}.

\begin{table}[h]
    \centering
    \small
    \begin{tabular}{lrlrl}
        \toprule
        & \multicolumn{2}{c}{ER}& \multicolumn{2}{c}{EE}\\
        Face Act & Frac & Coeff & Frac & Coeff\\\midrule
    
        HPos-&0.464& -0.036& 0.661 & { }0.003\\
        HPos+&0.646&{ }0.004& 0.605& -0.007\\
        HNeg+&0.753 &{ }\textbf{0.029*}&-&-\\
        HNeg-&0.709&{ }\textbf{0.023*}&0.744& { }\textbf{0.043***}\\
        SPos+&0.504&\textbf{-0.052***}&0.553&\textbf{-0.026*}\\
        SPos-&-&-&1.000& { }0.002\\
        Other&0.742&{ }\textbf{0.031***}&0.717&{ }\textbf{0.023*}\\
        SNeg+&-&-&0.435&\textbf{-0.038*}\\
        \bottomrule
    \end{tabular}
    \caption{Coefficients of the face acts, for ER and EE obtained from linear regression. A positive coefficient implies positive correlation. *, *** indicate statistical significance with p values $\leq$ 0.05 and 0.001.}
    \vspace{-0.7cm}
    \label{tab:correlation-donation-face-acts}
\end{table}

We observe several face acts which are statistically significant based on the corresponding p-value of the coefficients (highlighted in bold).


We observe unsurprisingly a positive correlation (0.029) for \textit{HNeg+} for ER since decreasing the imposition of the FTA is likely to influence donation. Likewise, a criticism of EE (\textit{HPos-}), decreases the likelihood of donation and thus has a negative correlation (-0.036). We also observe a negative coefficient for \textit{SPos+} (-0.052) which corroborates the finding of \citet{persuasion-for-good-2019}, that appealing to the credibility of the organization correlates negatively with donation outcome. 

Similarly, for EE, asserting one's face / independence (\textit{SNeg+}) corresponds to a decrease in the donation probability (-0.038). Likewise, \textit{HNeg+} corresponds to an increase in donation probability (0.043) since these face acts increase user engagement. We attribute the negative correlation for \textit{SPos+} (-0.026) and \textit{HPos+} (-0.007) to the fact that they often occur along with \textit{SNeg+} and hence decreases the outcome probability. Nevertheless, \textit{SPos+} and \textit{HPos+} increase the donation probability 60.5\% and 55.3\% of the times. However,  the  learned model  errs   in  assuming  that \textit{HPos-} (for EE) and \textit{HNeg-} (for ER) increases the donation probability and requires further investigation

We illustrate the effect of face act on the local donation probability through a  conversation snippet in Table \ref{Table: Annotated Conversation Snippet}. We observe a noticeable reduction in donation probability associated with \textit{SNeg+} (for EE) and \textit{HPos-} (for ER). Likewise, face acts corresponding to \textit{HNeg+} (for ER) and \textit{HPos+} (for EE) result in an increase in donation probability.

\section{Related Work}

Although politeness derailment and politeness evolution in dialogue have been previously investigated in the NLP literature \cite{ chang-danescu-niculescu-mizil-2019-trouble, danescu-niculescu-mizil-etal-2013-computational}, the prior work is distinguished from our own in that they do not explicitly model face changes of both parties over time. Rather, \citet{danescu-niculescu-mizil-etal-2013-computational} utilizes requests annotated for politeness to create a framework specifically to relate politeness and social power. Other previous work attempt to computationally model politeness, using politeness as a feature to identify conversations that appear to go awry in online discussions \cite{conversations-gone-awry}. Previous work has also explored indirect speech acts as potential sources of face-threatening acts through blame \cite{modeling-blame} and as face-saving acts in parliamentary debates \cite{face-saving}. 

The closest semblance of our work is with \citet{kluwer2011like, kluwer2015social}, which builds upon the notion of face provided by \citet{goffman1967interaction} and invents its own set of face acts specifically in the context of ``small-talk'' conversations. In contrast, our work specifically operationalizes the notion of the positive and negative face of \citet{brown1987politeness, brown1978universals}, which is well established in the Pragmatics literature and heavily acknowledged in the NLP community \cite{danescu-niculescu-mizil-etal-2013-computational, conversations-gone-awry, wang2012love, musi2018changemyview}. Moreover, we focus on analysing the effects of face acts in a ``goal-oriented'' task like persuasion, where there is an explicit threat or attack on face as opposed to small-talk scenarios, where the goal is building rapport or passing the time. Thus our work can be considered to be complementary to the prior work of \citet{kluwer2011like} and \citet{kluwer2015social}. 
It 
also enables us to draw insights from recent work in persuasion strategy to analyze face act exchanges in persuasion 
\cite{persuasion-for-good-2019, diyi-yang-persuasive}.




\section{Conclusion and Future Work}
In this paper, we present a generalized computational framework based on the notion of \emph{face} to operationalize face dynamics in conversations. We instantiate these face act exchanges in the context of persuasion and propose a dataset of $296$ conversations annotated with face acts. We develop computational models for predicting face acts as well as observe the impact of these predicted face acts on the donation outcome.

One important limitation of the current work is the assumption that all face acts have the same intensity/ranking. We seek to rectify this by separating the content and style of these face acts. We also wish to expand the current face framework to a more comprehensive politeness framework that incorporates notions of power and social distance between the interlocutors. We believe that our work may be extended to language generation in chatbots for producing more polite language to mediate face threats. Moreover, we intend to instantiate our proposed framework to other domains such as teacher/student conversations and other types of discourse such as social media narratives. 

\section{Acknowledgments}
We thank the anonymous EMNLP reviewers for their insightful comments. We are also grateful
to Xinru Yan, Meredith Riggs, and other members of the 
TELEDIA group at LTI, CMU. This research was funded in part by NSF Grants (IIS 1917668 and IIS 1822831) and from Dow Chemicals.


\bibliography{references}
\bibliographystyle{acl_natbib}
\newpage

\beginsupplement

\appendix
\section{Appendices}
\label{sec:appendix}


    

We present the hyper-parameters for all the experiments, their corresponding search space and their final values in Table \ref{tab:hyperparameters}.  We also present additional details of our experiments below.

\noindent(i) Each run took at most 1 hour on a single Nvidia GeForce GTX 1080 Ti GPU.


\noindent(ii) We present the number of parameters for our model in Table \ref{tab:num_params}.

\noindent(iii) All hyper-parameters were chosen based on the mean cross-validation performance.

\noindent(iv) Each validation split comprised 20\% of all Donor conversations and 20\% of all Non-Donor conversations, with the training split comprising the remaining 80\%. All validation splits are mutually exclusive and exhaustive. The data splits are made available in the supplementary material.

\begin{table}[h]
    \centering
    \scriptsize
    \begin{tabular}{l|c|c}
    \toprule
    Hyper-parameter & Search space & Final Value\\ 
    \midrule
    learning-rate (lr) & 1e-3, 1e-4& 1e-4 \\
    Batch-size & - & 1 conversation\\
    \#Epochs & 50, 100 & 50\\
    lr-decay & - & 0.966\\
    $d_{h1}$ & 300, 768 & 300\\
    $d_{h2}$ & 300 & 300\\
    $d_{fc}$ & 100 & 100\\
    $\alpha$ & [0, 0.25,0.5,0.75,0.9,1.0] & 0.75 \\
    $L_{D}$ & BCE, MSE & MSE\\
    Donation threshold & \{0.001 - 0.999\} & 0.813\\
     \bottomrule
    \end{tabular}
    \caption{Here we describe the search-space of all the hyper-parameters used in our experiments and describe the search space we used to find the hyper-parameters. All the experiments were run on a single 1080Ti GPU.  $d_{h1}$, $d_{h2}$ and $d_{fc}$ represents the hidden dimensions of Utterance GRU, Conversation GRU, and the Face act classifier. $\alpha$ is the hyper-parameter used to combine the face-act loss and donation loss denoted by $L_{f}$  and $L_{D}$ respectively.}
    \label{tab:hyperparameters}
\end{table}
\begin{table}[h]
    \centering
    \small
    \begin{tabular}{lrrr} \toprule
         Model& Parameter Size \\  \midrule
         BiGRU & 2.0M \\
         BiGRU-f & 2.1M\\
         BiGRU-sf & 2.2M \\
         HiGRU & 2.0M\\
         HiGRU-f & 2.1M \\
         HiGRU-sf & 2.4M \\
         BERT-BiGRU & 8.6M\\
         BERT-BiGRU-f &9.2M\\
         BERT-BiGRU-sf &10.4M\\
         BERT-HiGRU & 9.4M\\
         BERT-HiGRU-f & 10.2M\\
         BERT-HiGRU-sf & 11.5M\\
         \bottomrule
    \end{tabular}
    \caption{Number of parameters for each model in our experiments}
    \label{tab:num_params}
\end{table}

\begin{figure}[h]
    \centering
    \includegraphics[scale=0.5]{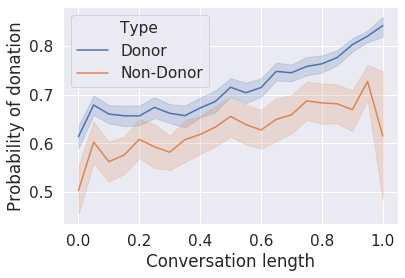}
    \caption{Plot of donation probability for Donor and Non-Donor conversations. We observe that the donation probability for Non-Donors at each step is lower than the corresponding probability for Donors.}
    \label{fig:conv-trend}
    \vspace{-0.5cm}
\end{figure}

We show how donation probability changes for both Donors and Non-Donors in Figure \ref{fig:conv-trend}. 

\begin{table*}[t]
    \centering
    \small
    \begin{tabular}{lll} \toprule
    Face act & Persuader (ER)  & Persuadee (EE)\\ \midrule
    SPos+ & (i) ER praises/promotes the good deeds of STC  &  (i) EE states her preference for other charities \\
    &(ii) ER shows her/ his involvement for STC & (ii) EE states that she does good deeds  \\  \hline
    HPos+ & (i) ER appreciates/praises EE’s
    &(i) EE shows willingness to donate\\
    &{ }{ }{ }{ }{ }generosity or time &{ }{ }{ }{ }{ }  to discuss the charity\\
    & (ii) Incentives EE to do a  good deed.  & (ii) EE acknowledges the efforts of STC. \\
    & (iii) Empathize/ agree with EE & (iii) Empathizes/ agrees with ER\\ \hline
    SPos- & 
    & (i) EE apologizes for not donating \\ \hline
    HPos- & (i)  ER criticizes EE & (i) EE doubts/ questions STC or EE\\
    & & (ii) EE is not aware of STC\\ \hline
    SNeg+ & & (i) Rejects donation out-right\\
    & & (ii) Cites reason for not donating at all or\\
    && { }{ }{ }{ }{ }{ }not donating more. \\ \hline
    HNeg+ & (i) ER provides EE convenient ways to donate. & \\
        & (ii) ER apologizes for inconvenience/ intrusion. &\\
        & (iii) ER decreases the amount of donation.& \\ \hline
        
    HNeg- & (i) ER ask’s EE’s time/ permission for discussion.  & (i) EE asks ER questions about STC. \\
    & (ii) ER asks EE for donation. & \\
    & (iii) ER asks EE to donate more.& \\ 
    \bottomrule
    \end{tabular}
    \caption{Instantiating predicates corresponding to the different face acts in the context of persuasion.}
    \label{tab:predicates }
\end{table*}

\begin{figure*}[h]
    \centering
    \vspace{-2.5cm}
    \includegraphics[scale=0.3]{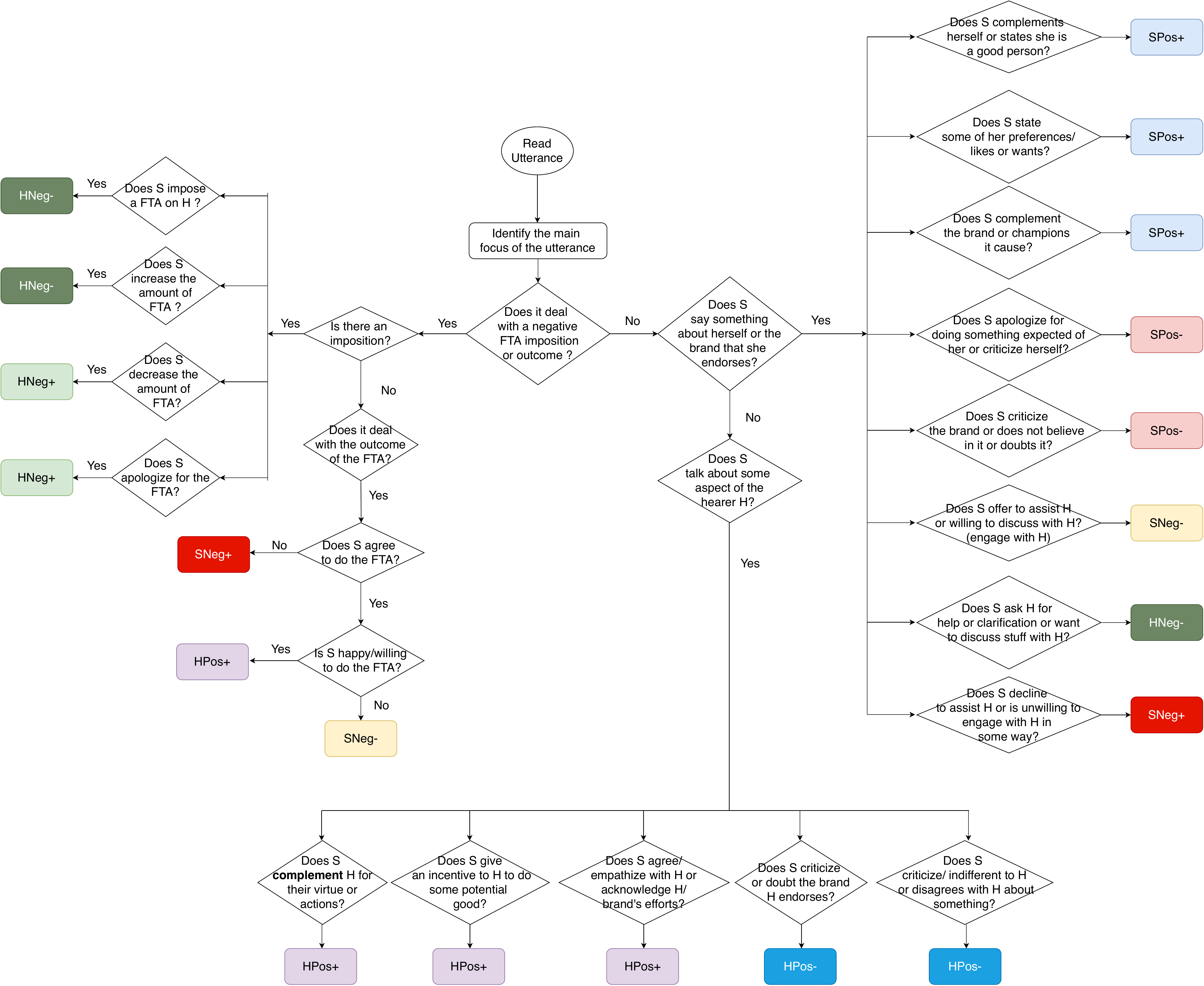}
    \caption{Flowchart outlining the annotation framework employed for labeling face acts in persuasion conversations.}
    \label{fig:my_label}
\end{figure*}


\end{document}